\documentclass[10pt,a4paper,twocolumn]{article}
\usepackage[T1]{fontenc}
\usepackage[utf8]{inputenc}
\usepackage[british]{babel}
\usepackage{lmodern}
\usepackage[margin=0.78in,columnsep=0.3in]{geometry}
\usepackage{amsmath,amssymb}
\usepackage{graphicx}
\usepackage{booktabs}
\usepackage{array}
\usepackage{caption}
\usepackage[round,authoryear]{natbib}
\usepackage[hidelinks]{hyperref}
\bibpunct{(}{)}{;}{a}{,}{,}

\usepackage{stfloats}
\begin{document}

\twocolumn[{%
\begin{center}
{\LARGE\bfseries Compaction, decay and reinforcement in a two-tier conversational memory: a paired evaluation against retrieval-augmented generation\par}
\vspace{0.9em}
{\large Osamudiamen Edogun\textsuperscript{1}, Joshua Obigwe\textsuperscript{1}, Irfan Mehmood\textsuperscript{2}\par}
\vspace{0.2em}
{\normalsize \textsuperscript{1}Independent researcher\quad \textsuperscript{2}University of Bradford \quad \texttt{edogunosamudiamen@gmail.com, joshuaobigwe@gmail.com, I.Mehmood4@bradford.ac.uk}\par}
\vspace{0.2em}
{\normalsize 9 October 2026\par}
\end{center}
\vspace{0.6em}
\begin{center}
\begin{minipage}{0.86\textwidth}
\small
{\centering\textsc{Abstract}\par}
\vspace{0.3em}
A conversational agent that remembers has to decide what to keep. Retrieval-augmented generation keeps the transcript and searches it; the alternative is to compact each turn into atomic facts and search those. We compare the two on LoCoMo \citep{maharana2024}, with both arms answering all 1,985 questions in one window, against the same stores, with the same generator and judge, and both predictions on one row, which licenses McNemar's exact test on the 512 questions where they disagree \citep{mcnemar1947,dietterich1998}.

On the four categories comparable work reports, compaction scores 71.4 per cent against 64.6, a lead of 6.8 points (p = 5.5e-08) at 34 per cent fewer prompt tokens. On all five, including the adversarial set comparable work drops, the lead falls to 2.0 and stops being significant, because the baseline leads adversarial by 14.6.

That loss has a structure. Our lead falls as the reference answer demands more source wording, 9.6 points ahead at one or two reference words and 3.1 behind past eight, with token overlap moving the same way, so the judge is not the cause. Compaction keeps what a fact asserts and discards how it was said, so it wins where the answer is a value and loses where it is a phrasing. We also report four negative results with their power calculations.

\end{minipage}
\end{center}
\vspace{1.4em}
}]

\section{Introduction}

An assistant that holds a conversation over months faces a storage question before a retrieval question. The transcript grows without bound, so something has to be dropped, summarised or indexed, and that choice determines what can later be answered.

Retrieval-augmented generation keeps everything and searches it at query time \citep{lewis2020,karpukhin2020}. Its failure modes are documented, since long contexts degrade in the middle \citep{liu2024} and passages that are topically close but factually irrelevant can hurt more than they help \citep{cuconasu2024}.

The alternative is to compact. An extractor turns each turn into atomic assertions, which are deduplicated and linked, and retrieval runs over those rather than over the text \citep{packer2023,zhong2024,rasmussen2025,xu2025,chhikara2025}. Compaction is cheap, since an assertion is shorter than its turn, and lossy in a way that is hard to measure, because the extractor decides what counts as content out of sight of the query.

This paper measures the trade. The system under test, NMAFC (Neuromorphic Memory Architecture for Conversational AI), stores facts carrying a weight that decays exponentially in elapsed turns and is restored on retrieval, following the forgetting curve \citep{murre2015} and the long-term potentiation literature \citep{bliss1973}. A fact whose weight crosses a threshold leaves the working tier for a searchable archive. Retrieval fuses a dense and a lexical list by reciprocal rank \citep{cormack2009} and hydrates the top facts with their source turns.

The comparison itself is the part we want to defend. Published comparisons of memory systems commonly run each arm separately and subtract the totals, which cannot support a paired test and here discards most of the resolution: the arms agree on 1,473 of 1,985 questions, so a test on the totals tests the wrong quantity. We ran both arms in one window against the same persisted stores, with the same generator and judge, and wrote both predictions to one row, so every comparison below is McNemar's exact test on the discordant pairs.

We contribute the protocol, a measured account of what compaction buys and costs at 34 per cent lower context spend, the finding that the advantage falls with reference-answer length and reverses past eight words, four negative results with the power calculations that make them negative rather than merely unobserved, and a correction to a latency claim we made from unpaired runs. The per-question predictions of both arms are released, so every figure here can be recomputed without an API key.

\section{Related work}

\subsection{Retrieval over transcripts}

Dense retrieval over passages, trained with a dual encoder, established the pattern that most conversational memory baselines still follow \citep{karpukhin2020,reimers2019}, and retrieval-augmented generation joined that retriever to a generator \citep{lewis2020}. Practical systems add a lexical channel, usually BM25 \citep{robertson2009}, because dense and sparse retrieval fail on different queries; we fuse the two lists by reciprocal rank because it needs no score calibration between channels \citep{cormack2009}. The vector index is a solved engineering problem at our scale \citep{malkov2020,johnson2021}.

The limits are documented. Accuracy depends on where in a long context the relevant passage sits, with a pronounced dip in the middle \citep{liu2024}, and topically related passages that do not contain the answer degrade output quality \citep{cuconasu2024}. Both argue for placing less in the prompt rather than more, which is the argument compaction makes on its own behalf.

\subsection{Agent memory and deliberate forgetting}

MemGPT pages between tiers of a memory hierarchy under the model's control \citep{packer2023}, MemoryBank adds a forgetting curve and updates memories with use \citep{zhong2024}, Zep gives facts validity intervals in a temporal knowledge graph so they can be superseded rather than overwritten \citep{rasmussen2025}, A-MEM links memories into a network at write time \citep{xu2025}, and Mem0 targets production with an extraction and consolidation pipeline \citep{chhikara2025}. Ours shares decay with MemoryBank, validity intervals with Zep and two tiers with MemGPT. What they have in common with ours is that their published numbers come from separately-run arms, which is the gap Section 4.2 addresses.

Deliberate forgetting is older than the systems that implement it. Ebbinghaus's retention curve, replicated a century later, is close to exponential in elapsed time \citep{murre2015}, and long-term potentiation supplies the complementary mechanism, since repeated activation strengthens a pathway durably \citep{bliss1973} and consolidation moves a memory from a labile to a stable state \citep{squire2015}. Spreading activation is the retrieval-side counterpart \citep{collins1975,anderson1983}. Catastrophic interference is a different failure with a similar name \citep{mccloskey1989,kirkpatrick2017}: those systems forget because new learning overwrites old weights, ours because a policy decided to.

We state the limit of the analogy rather than gloss it. Work on synaptic engrams reports memory surviving the elimination of most synapses, with clustered connectivity rather than raw synapse count carrying it \citep{lin2026}, and mapping spatial clustering on a dendrite onto graph clustering over entity names is an analogy, which is why our implementation carries it as an ablation switch that is off by default.

\subsection{Evaluating conversational memory}

LoCoMo supplies very long synthetic conversations with questions categorised by the reasoning each needs \citep{maharana2024}, and LongMemEval covers a related setting with a different taxonomy \citep{wu2024}. Both are scored by a model judge, a practice whose agreement with human raters, and whose position and verbosity biases, have been characterised \citep{zheng2023}. Token-overlap F1 against a reference, our secondary measure in Section 5.3, comes from extractive question answering \citep{rajpurkar2016}. The hallucination survey of Huang et al. gives the wider framing for why grounded context is worth paying for \citep{huang2025}.

\section{Method}

\subsection{Overview}

\begin{figure*}[tb]
\centering
\includegraphics[width=0.70\textwidth]{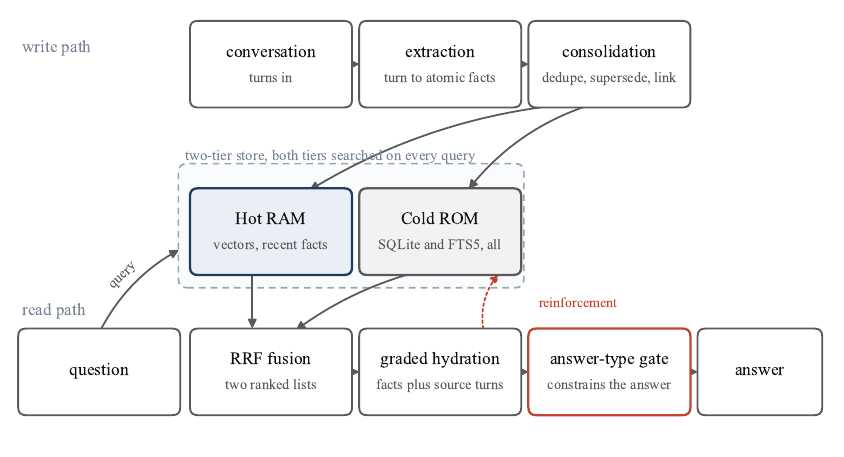}
\caption{The two paths through the system. Write path across the top, read path across the bottom, the two storage tiers in the middle. The dotted return arrow is reinforcement: reading a fact rewrites it.}\label{figure:arch}
\end{figure*}

Figure \ref{figure:arch} shows the system. Writing runs extraction into consolidation into two tiers, a working tier of dense vectors and an archive in SQLite with a full-text index. Reading searches both tiers, fuses the ranked lists, renders the survivors into the prompt with the raw turns behind the best of them, and reinforces every fact it returned. Appendix A lists every parameter that differs from zero or from a library default.

\subsection{Extraction and consolidation}

The extractor is a language model with a tool-calling schema that forces each record to name an entity and assert one thing about it, with a validity date, a speaker attribution and a list of related entities. Consolidation deduplicates against what is stored and marks superseded facts invalid rather than deleting them. Over the ten LoCoMo conversations the extractor produced 8,276 facts and 11,902 links.

Two properties bear on the results. It resolves relative dates, so a turn saying ''two weekends ago'' becomes a fact carrying an absolute date, which Section 5.2 shows is where the temporal advantage comes from. It also drops qualifiers: a turn saying a novel was gripping yields a fact that the speaker read the novel, and the word ''gripping'' is gone. In Tulving's terms it converts an episodic record, carrying the occasion and its wording, into a semantic one carrying the proposition \citep{tulving1985}, and that conversion is the subject of Section 5.3.

Link resolution matters more than it looks. Of the 11,902 links written, 38.8 per cent named an entity that was never stored, most of them near misses on a compound name, so link targets are resolved by token overlap at a threshold of 0.5 and what matches nothing is discarded.

\subsection{Decay}

Each mutable fact carries a weight $w \in (0, 1]$, initialised at 1, which falls exponentially in the number of conversation turns since the fact was last reinforced:

\begin{equation}\label{equation:decay}
w(\Delta t) = w_0 \, e^{-\lambda_{\mathrm{eff}} \Delta t}
\end{equation}

where $\Delta t$ is measured in turns rather than in wall-clock time, because a conversation's rate of arrival is not the memory's business. The effective rate is a product of three terms:

\begin{equation}\label{equation:lambda}
\lambda_{\mathrm{eff}} = \lambda_{0}(\tau) \cdot e^{-\eta k} \cdot (1 - \beta C)
\end{equation}

The first is a base rate set by the fact's type $\tau$, with $\lambda_0$ of 0 for a core anchor, 0.05 for active context and 0.69 for ephemeral state. A core anchor never decaying is a policy decision rather than a measurement: an assertion about who the user is should not expire because it went unmentioned.

The second is consolidation, where $k$ counts retrievals and $\eta = 0.15$ sets how much each one flattens the curve, so a fact retrieved ten times decays at $e^{-1.5} \approx 0.22$ of its base rate \citep{bliss1973}. An active-context fact never retrieved crosses the pruning threshold after 46 turns; retrieved ten times it survives 206. Figure \ref{figure:decay} in Appendix A plots this.

The third is clustering protection, where $C$ is the local clustering coefficient of the fact's entity and $\beta$ scales it. We set $\beta = 0$, removing the term; it exists as an ablation control and because the synaptic evidence points at clustered rather than dense connectivity \citep{lin2026}. No result here uses it.

A fact is pruned from the working tier when $w < w_{\mathrm{prune}} = 0.1$. Pruning demotes rather than deletes, so the fact stays searchable in the archive.

\subsection{Reinforcement}

Every fact returned by a retrieval is reinforced: $w \leftarrow 1$, $k$ increments, and the last-reinforced turn advances. Reinforcement is buffered and written back on flush rather than per query, because the write path rewrites each record as a delete followed by an add, and doing that per query makes retrieval latency grow with the number of queries served. A buffered reinforcement is invisible to any retrieval before the flush, so if ranking read weight a deferred run could rank differently from an immediate one; our default ranking does not read weight, for the reason in Section 5.6.

The rule also means that reading the memory writes to it. An evaluation asking 1,985 questions mutates the store, so the stored values of $k$ afterwards measure the benchmark rather than the deployment. Section 4.2 describes how the harness handles this.

\subsection{Storage, fusion and ranking}

The working tier holds facts as dense vectors with metadata. The archive holds every fact ever written, including the superseded and the pruned, in SQLite with an FTS5 index and a second dense index. An early design searched the archive only when the working tier returned nothing convincing; we measured that gate and removed it, so both tiers are now searched on every query.

Each retrieval produces up to five ranked lists: dense and breadth-first graph expansion over the working tier, and dense, lexical and graph expansion over the archive. Reciprocal rank fusion combines them \citep{cormack2009}. For a fact $d$ at rank $r_i(d)$ in list $i$:

\begin{equation}\label{equation:rrf}
s(d) = \sum_{i} \frac{1}{k_{\mathrm{rrf}} + r_i(d)} + \omega \, g(d)
\end{equation}

with $k_{\mathrm{rrf}} = 60$ and the top 20 facts by $s$ surviving into the prompt. The rule needs no calibration between channels, which is why it is preferable to a weighted score sum when one channel returns cosine similarities and another BM25 scores.

The second term is a grounding bonus, where $g(d) \in [0, 1]$ measures how well the question's words match the raw turn the fact came from. It is additive rather than a sixth list because adjacent ranks inside one list differ by about 0.0003, so a weight of that order separates facts the other channels rank identically and leaves everything that won on merit where it was. A recency boost and a weight signal $\omega_w w(d)$ also exist and are zero by default; Section 5.6 reports the measurement behind the second.

\subsection{Graded hydration}

Extraction is lossy, so the prompt carries the raw turns behind the top-ranked facts alongside the facts themselves. Let $F$ be the ranked fact list and $T(f)$ the source turn of fact $f$. The rendered context is

\begin{equation}\label{equation:hydration}
\mathcal{C} \;=\; \bigcup_{j \leq m} \mathrm{render}(F_j) \;\cup\; \bigcup_{j \leq h} \mathrm{trunc}_{\ell}(T(F_j))
\end{equation}

The first union is the fact list and the second the hydrated turns, with $m$ facts rendered, $h$ hydrated and $\ell$ a line budget per turn. The default is $m = 20$, $h = 5$ and no truncation. The halves are not equally priced: widening $h$ from five to ten adds 331 tokens, so a hydrated turn costs about 66 tokens and a rendered fact about 32. Section 5.4 measures which half is worth the money.

\subsection{The answer-type gate}

Some questions name the category their answer must belong to. ''Which state does she live in'' demands a state, and ''Stamford'' is wrong even though Stamford is where she lives. A deterministic function of the question text detects these and appends a rule to the prompt when it fires, which it does on 319 of 1,985 questions here. Accuracy on those is 61.4 per cent against 63.8 where the gate stays quiet, so it does not lift the questions it touches to the level of the others. We report it as shipped and unproven, and Appendix E gives the measurement it was built on.

\section{Experimental setup}

\subsection{Benchmark and baseline}

LoCoMo supplies ten synthetic multi-session conversations with question-answer pairs \citep{maharana2024}. We answer 1,985 questions: 841 open-domain, 446 adversarial, 321 temporal, 281 single-hop and 96 multi-hop, with per-conversation counts from 105 to 259. Comparable published work reports the four we call scored and drops adversarial. We report both totals throughout, because a category is not excludable on the grounds that we lose it.

The adversarial category needs a description, because its name misleads here. These are not unanswerable questions. Their reference answers are ordinary open-ended statements, frequently long paraphrases of a turn: ''self-care is important'', ''creating a family for kids who need one''. They are a wordiness test rather than a refusal test, which is what Section 5.3 exploits.

The baseline is a retrieval-augmented generation arm over the same conversations. It chunks the transcript, embeds the chunks with the embedding model our arm uses, retrieves by similarity, and passes the retrieved text to the same generator with the same answer-format instruction, running no extraction, decay, reinforcement or fusion. It is strong rather than straw: it wins one of the five categories outright and ties a second.

\subsection{The paired protocol}

Both arms answered every question inside a single window. For each question the harness retrieved with our arm, generated, retrieved with the baseline, generated, and wrote both predictions, both context sizes and both latencies to one row, which the judge then scored.

Three details make the pairing sound. Stores are never written in place: the benchmark opens them through a read-only wrapper that clears the pending-reinforcement buffer before closing, so a pass leaves the store byte-identical. Without this, question 900 would be answered against a store that questions 1 to 899 had modified, and a rerun would not reproduce. Both arms see the same store state, since they are invoked back to back on one question, so there is no drift in model version, provider throttling or index state between them; throttling was recorded per call and was zero for both. And both predictions are judged in the same call sequence with the arms unlabelled, because a judge that scores one arm on Monday and the other on Thursday is measuring the judge as much as the arms \citep{zheng2023}.

\subsection{Scoring and statistics}

A model judge marks each prediction against the reference as correct or incorrect. Judge-based scoring is standard on this benchmark and carries known biases, verbosity among them \citep{zheng2023}. Three choices reduce the exposure. Both arms receive the same answer-format instruction, so neither is systematically more verbose: median prediction length is 20 characters for both and the ninetieth percentile 50 against 51. Both are judged in the same call sequence with the arms unlabelled. And Section 5.3 repeats the central analysis with token-overlap F1 \citep{rajpurkar2016}, which has no model in the loop.

Accuracy differences are tested with McNemar's exact test, two-sided, on the discordant pairs \citep{mcnemar1947}, following the standard argument for comparing two learners on one sample \citep{dietterich1998}: where the arms agree 1,473 times out of 1,985, the 512 disagreements are the entire evidence. We report the exact binomial tail rather than the chi-squared approximation, because several categories have small discordant counts where the approximation is poor, multi-hop with 18 being the extreme. Interval estimates are Wilson score intervals \citep{wilson1927}, which behave near 0 and 1 where the Wald interval does not.

We also report the minimum detectable effect, because four results here are negative and a negative result without a power statement is not a result. At the discordance rate of 0.25 observed here, the smallest lead a two-sided paired test resolves at $\alpha = 0.05$ is 9.4 points on 108 questions, 5.5 on 322 and 2.2 on 1,985. Every effect in this paper except the temporal one is smaller than nine points, which is why a sample size is quoted with each. Appendix B gives the derivation and the full ladder.

All model calls run through one hosted endpoint, an Azure AI Foundry deployment: \texttt{DeepSeek-V4-Pro} serves extraction and generation in both arms, OpenAI's \texttt{text-embedding-3-small} (1,536 dimensions) serves embeddings in both, and \texttt{Kimi-K2.6} serves as judge — chosen because it is the only non-DeepSeek model family the endpoint serves, so the model being scored is not scoring itself. All runs were executed between August and September 2026 against these deployments. The endpoint sits behind a batching proxy, so outputs vary between identical calls even at temperature zero; the released per-question predictions are therefore the record of the runs, not any rerun of them. Retrieved-context size is in tokens, approximated as characters divided by four; the approximation is crude and applied identically to both arms, so what is claimed is the ratio rather than the absolute.

\section{Results}

\subsection{The main comparison}

Table \ref{table:main} gives the paired result, which Figure \ref{figure:acc} in Appendix C plots.

\begin{table*}[tb]
\centering
\caption{Paired LoCoMo result, 1,985 questions, both arms in one window against the same stores. Intervals are Wilson 95 per cent. Discordant counts are questions won and lost. Context is mean retrieved tokens.}\label{table:main}
\scriptsize
\begin{tabular}{lrrrrrrr}
\toprule
Category & n & NMAFC & RAG & Lead & Won/lost & McNemar p & Context \\
\midrule
single-hop & 281 & 50.9 [45.1, 56.7] & 45.6 [39.8, 51.4] & +5.3 & 51/36 & 0.133 & 980 vs 1,486 \\
temporal & 321 & 71.7 [66.5, 76.3] & 46.1 [40.7, 51.6] & +25.5 & 101/19 & 1.1e-14 & 943 vs 1,422 \\
multi-hop & 96 & 57.3 [47.3, 66.7] & 44.8 [35.2, 54.7] & +12.5 & 15/3 & 0.0075 & 961 vs 1,410 \\
open-domain & 841 & 79.8 [76.9, 82.4] & 80.3 [77.4, 82.8] & -0.5 & 71/75 & 0.804 & 967 vs 1,461 \\
adversarial & 446 & 35.9 [31.6, 40.4] & 50.4 [45.8, 55.1] & -14.6 & 38/103 & 4.1e-08 & 960 vs 1,486 \\
scored (4) & 1,539 & 71.4 [69.1, 73.6] & 64.6 [62.2, 66.9] & +6.8 & 238/133 & 5.5e-08 & 964 vs 1,454 \\
all (5) & 1,985 & 63.4 [61.3, 65.5] & 61.4 [59.2, 63.5] & +2.0 & 276/236 & 0.085 & 963 vs 1,461 \\
\bottomrule
\end{tabular}
\end{table*}

Three readings of this table are all correct, and a paper reporting one of them would be misleading. On the four categories comparable work reports, we lead by 6.8 points at 34 per cent lower context spend, and the result is not close to the significance boundary. On all five, we lead by 2.0 points, which the test does not distinguish from zero. On the fifth category alone, we lose by 14.6 with 103 questions lost against 38 won.

The pairing is what makes those readings available. The scored-four totals differ by 105 questions, which is what a subtraction would see; the paired test sees 371 discordant questions and the direction of each. Appendix B breaks the discordance down by category and shows that no single conversation carries the result: our per-conversation accuracy runs from 56.4 to 73.2 per cent and the baseline's from 55.2 to 71.1.

\subsection{Where the advantage comes from}

The temporal category carries the result. A lead of 25.5 points on 321 questions, with 101 won and 19 lost, is larger than the entire scored-four lead and survives any reasonable correction for multiple comparisons.

The cause is in the extractor rather than the retrieval. LoCoMo temporal questions are phrased relatively: ''two weekends before 17 July 2023'', ''the month after she started the new job''. The baseline retrieves a chunk containing the relative phrase and asks the generator to do the arithmetic at answer time, on text that may not contain the anchor date. Our extractor does it at write time, once, with the session date in hand, and stores an absolute date, so the advantage is bought when the fact is written and collected at query time.

The failure mode is visible in the baseline's own predictions. Asked when a speaker painted a sunrise, it answers ''last year''; when she plans to go camping, ''next month''; when a picnic happened, ''yesterday''. The retrieved chunk holds the phrase and not the anchor, so the deixis survives into the answer, where our arm answers 2022, June 2023, and late June or early July 2023. Appendix D lists these verbatim.

Multi-hop shows the same shape more weakly, at +12.5 points on 96 questions with 15 won and 3 lost, where the small discordant count is why the exact test is the right one to quote. Single-hop, at +5.3 with 51 won and 36 lost, does not reach significance on 281 questions, where the minimum detectable effect is about 5.9 points, so it is an effect we are not powered to resolve rather than one shown to be absent. Open-domain is a tie at 79.8 against 80.3, p = 0.804, at 34 per cent less context.

\subsection{Where the loss is, and what it is made of}

The adversarial category is the loss, and its structure is the most useful finding in this paper.

Our first reading was that it is a refusal problem, our arm declining where the baseline attempts an answer. Measured, that is a small part of it: over the 446 adversarial questions our arm declines on 16.6 per cent and the baseline on 16.8, and of the 103 questions the baseline wins, 18 are ours declining. The rule that counts a decline is in the released analysis script and should be read before its output is believed, since two different rules put the same phenomenon at 3.6 per cent and at 16.6.

The other 85 losses are the interesting ones, because on those our arm answered confidently and the judge rejected it. Token overlap with the reference on those 85 is 0.129 for us and 0.765 for the baseline. We are not failing to answer; we are answering in different words.

\begin{figure*}[tb]
\centering
\includegraphics[width=0.70\textwidth]{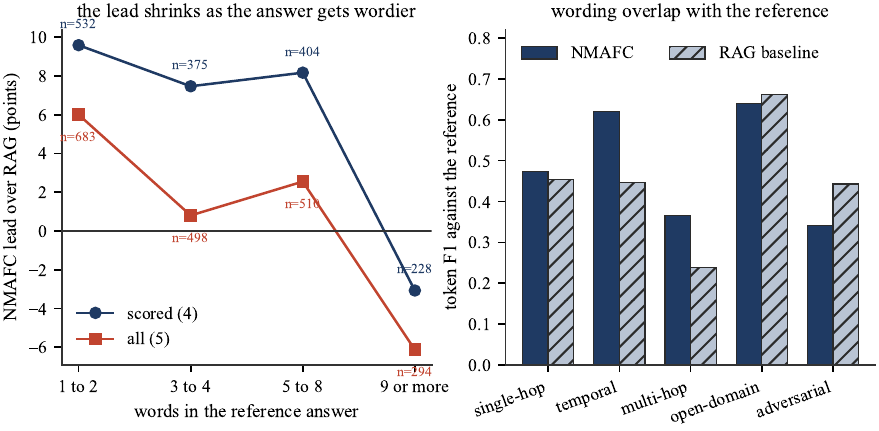}
\caption{Left, the paired lead by reference-answer length, on all five categories and on the scored four. Right, mean token-overlap F1 against the reference by category, which has no model in the loop.}\label{figure:goldlen}
\end{figure*}

The structure of that failure is in the length of the reference answer. Table \ref{table:goldlen} and Figure \ref{figure:goldlen} split all 1,985 questions by how many words the reference contains.

\begin{table*}[tb]
\centering
\caption{The paired lead as a function of reference-answer length. The split is on the reference, so it is fixed before either arm runs.}\label{table:goldlen}
\scriptsize
\begin{tabular}{lrrrrrr}
\toprule
Reference length & n (all 5) & Lead, all 5 & p & n (scored 4) & Lead, scored 4 & p \\
\midrule
1 to 2 words & 683 & +6.0 & 0.0027 & 532 & +9.6 & 1.2e-05 \\
3 to 4 words & 498 & +0.8 & 0.781 & 375 & +7.5 & 0.0023 \\
5 to 8 words & 510 & +2.5 & 0.305 & 404 & +8.2 & 0.0017 \\
9 or more words & 294 & -6.1 & 0.057 & 228 & -3.1 & 0.410 \\
\bottomrule
\end{tabular}
\end{table*}

The lead falls as the reference gets wordier, and on the scored four it crosses zero somewhere past eight words. Inside the adversarial category alone the gradient is steeper: at one or two reference words we score 51.0 per cent against 57.6, and at nine or more we score 19.7 against 36.4.

The right panel of Figure \ref{figure:goldlen} gives an independent check with no judge in it. Mean token-overlap F1 \citep{rajpurkar2016} favours our arm on temporal (0.620 against 0.446) and multi-hop (0.365 against 0.239), ties on single-hop, and favours the baseline on adversarial (0.340 against 0.443). The judge and the overlap metric agree, so the pattern is not a judging artefact.

The mechanism follows from what compaction does. Extraction preserves the assertion and discards the phrasing, so when the reference is a value, a date, a name, a place, our representation holds exactly that value while the baseline has to find it inside a chunk; when the reference is a phrasing, the baseline holds it verbatim and can echo it while our arm reconstructs it from a summary that deleted the words. Asked what a speaker was excited about in her adoption process, against a reference of ''creating a family for kids who need one'', the baseline answers ''making a family for kids who need one'' and we answer ''inspired by her friend's positive experience''. Both describe the same enthusiasm; one shares six words with the reference. Appendix D gives eleven further cases, the first the filter returned in each direction rather than illustrations chosen after reading them.

This is a property of the representation, and it is not fixable in the prompt. Section 5.6 reports two attempts.

\subsection{Context economy}

Our arm places 963 tokens of retrieved context in the prompt against the baseline's 1,461, a reduction of 34 per cent, and the reduction is uniform across categories. Figure \ref{figure:pareto} in Appendix C plots accuracy against context spend.

Its left panel is an ablation on retrieval width, run paired over all 841 open-domain questions. The shipped setting renders facts and five hydrated turns at 967 tokens for 79.7 per cent. Adding the grounding term of Equation \ref{equation:rrf} at $\omega = 0.003$ buys 0.8 points for 78 more tokens (31 won, 24 lost, p = 0.419), and widening hydration to ten turns as well buys 1.7 points for 331 more tokens (36 won, 22 lost, p = 0.087). The baseline sits at 80.3 per cent for 1,461 tokens.

Two things follow. Extra context does buy accuracy here, and the purchase is not significant at either width we tested. And a hydrated turn costs about twice what a fact costs, so buying accuracy through hydration is the expensive route: passing the baseline takes 1,298 tokens, where the fact list alone lands within 0.6 of a point of it at 967. We did not ship the wide setting, because the category of an incoming question is not knowable at inference time, so a setting that helps open-domain would be paid for on every question, including the 446 adversarial ones where it does not help.

\subsection{Latency}

We previously reported, from two separately-run passes, that our arm was several times slower than the baseline. That claim came from comparing means across runs, and it was wrong. Measured paired, in milliseconds by nearest rank, our arm runs 1,741 at p50, 2,407 at p75, 3,570 at p90, 28,348 at p99 and 93,778 at maximum, for a mean of 2,721; the baseline runs 1,769, 2,398, 3,360, 12,149 and 93,848, for a mean of 2,295. Appendix C plots the distributions.

The arms are indistinguishable through p90, our median being 28 milliseconds faster and our p90 210 milliseconds slower. The 19 per cent gap in means is entirely tail, and both maxima near 93.8 seconds are the provider's retry ladder rather than either system's work. Reporting the mean of a distribution with that tail was the original error.

\subsection{Negative results}

Four mechanisms were built, measured, and either switched off or left unwired. Each cost a paid benchmarking run, and each is reported with the sample size that makes it a negative result rather than an absence of evidence. Appendix E gives the counts and p-values.

Ranking by decay weight, which is what having a decay curve was for, costs 5.3 points of answer reachability when $\omega_w w(d)$ is added to Equation \ref{equation:rrf}. The curve therefore earns its place by controlling what is stored rather than by ordering what is retrieved. The archival search gate, which consulted the archive only when the working tier's best score fell below a threshold, never usefully fires, and widening the archive's budget behind it hurts recall, so it was removed.

A deterministic list-shape gate, measured paired on the 214 questions where it fires, scored 36.4 per cent against the shipped setting's 36.9 at 1,286 tokens against 968, p = 1.0. It ships as a library function that nothing calls. An earlier reading on 108 questions had looked worth +5.6 points, where the minimum detectable effect is 9.4, so it was nine coin flips.

Two answer-shape clauses in the prompt, both aimed at the loss of Section 5.3, cancelled. A short-phrase instruction gained 5.4 points on adversarial and lost 4.0 on temporal, for 0.0 across 1,984 questions; an exact-wording instruction gained 0.2 and lost 3.4 across 767. The clause that recovers wordiness costs precision at close to the same rate.

\section{Discussion}

\subsection{What the trade actually is}

Compaction and transcript retrieval are not competing implementations of one idea. They store different things, and the difference shows up as a systematic split by question type. Compaction wins where the answer is a value the extractor can compute or normalise once and store, and temporal arithmetic is the clearest case at +25.5 points, won at write time. Transcript retrieval wins where the answer is a phrasing, because it still has the phrasing and we do not. The gradient in Table \ref{table:goldlen} measures that boundary directly.

A practitioner can use it. A support agent answering ''when did the contract renew'' and ''which plan is the customer on'' is on the left of Figure \ref{figure:goldlen}, where compaction is worth 6 to 10 points at a third less context. One answering ''what did the customer say they were unhappy about'' is on the right, where the transcript wins. The two are composable, and our hydration mechanism is a partial composition already; the reason we did not widen it is that the question's category is not knowable at inference, so a routing decision would have to be made on the question text alone.

\subsection{What the pairing changed}

The protocol of Section 4.2 was not free, and it changed two conclusions. It corrected the latency claim, where two unpaired runs had produced means differing by a factor of several and the paired measurement shows the arms indistinguishable through p90; the unpaired comparison was measuring provider conditions on two different days.

It also made the negative results interpretable. Three of the four mechanisms in Section 5.6 look positive at small sample sizes, and the list gate looked worth +5.6 points on 108 questions. Only a design that puts both arms on the same 214 questions, and counts wins and losses rather than totals, exposes that as noise. The rule we take from it is to avoid sizing an experiment on this benchmark below about 800 questions unless the effect sought is larger than three points.

\subsection{On judge-based scoring}

Section 4.3 lists the three things that limit our exposure to a biased judge. What none of them rules out is that the judge's notion of a correct paraphrase differs systematically from a human reader's on the long-reference questions where the baseline wins. That would change the size of the adversarial gap, but not its direction, because the overlap metric moves the same way.

\section*{Limitations}

Construct validity. LoCoMo is synthetic, so it may not reproduce the interruptions, corrections and topic drift of a real transcript. Its adversarial references are long paraphrases, and a benchmark whose references were short values would report a larger lead for us on exactly the same system.

Internal validity. Both arms share a generator, an embedding model and a judge, so a defect in any of the three is common-mode and cancels in the paired comparison. It does not cancel in the absolute numbers, which should not be compared against published figures obtained with different models.

External validity. Ten conversations is a small population, though 1,985 questions is not a small sample. The per-conversation spread is 17 points wide for our arm and 16 for the baseline, so a study on one conversation could report almost anything.

Statistical validity. We run seven category-level tests plus two totals without a family-wise correction. Three survive any correction that is not absurd: temporal, adversarial and the scored-four total, all at p below 1e-07. Single-hop at 0.133 and the all-five total at 0.085 are reported as undetermined at this sample size.

Reproducibility and scope. Model outputs are not deterministic, and cross-run drift here, measured on a category-level comparison repeated a week apart, was 0.040 in accuracy, so effects smaller than about four points should not be trusted across runs even where the sample size resolves them within one. Ingesting the ten conversations costs several hours of paid inference, so the ablations of Section 5.6 ran against already-persisted stores; any effect that would have changed what was written at ingestion time is outside what they can see.

\section{Conclusion}

We compared a compacting two-tier conversational memory against transcript retrieval on 1,985 LoCoMo questions, both arms in one window against the same stores, both predictions on one row. Compaction leads by 6.8 points on the four categories comparable work reports, at 34 per cent less retrieved context, and by 2.0 across all five, which the test does not separate from zero.

The category we lose has a structure that generalises past this benchmark. Our advantage falls as the reference demands more source wording, 9.6 points ahead at one or two words and 3.1 behind past eight, with token overlap moving the same way. Compaction keeps what was asserted and discards how it was said, and that is the whole trade.

Two directions follow. A router predicting reference-answer shape from the question alone could pay for hydration only where the gradient says it is worth it, and Section 5.4 gives the price. An extractor preserving a short verbatim span alongside each assertion would attack the loss at its source rather than at the prompt, where Section 5.6 shows it cannot be attacked.

The protocol is the part we would most like to see adopted: two arms, one window, one row per question, a test on the discordant pairs. It costs one benchmarking run instead of two, and it is the difference between a result and a pair of numbers.

\section*{Data and code availability}

The per-question predictions of both arms, for every run quoted here, are released at https://github.com/BigBrown10/nmafc-paper, together with every script and JSON file listed in Appendix F. Every number in every table can be recomputed with \texttt{python analysis.py} and every figure with \texttt{python make\_figures.py}, using the standard library, matplotlib and the released JSON. Neither script calls a model or requires an API key.

\section*{Appendix A. Configuration and decay}

\begin{table}[tb]
\centering
\caption{Shipped configuration. Every parameter that differs from zero or from a library default is listed. Values marked as ablation controls are inactive at their defaults and exist so that the corresponding mechanism can be switched off and attributed.}\label{table:config}
\small
\begin{tabular}{p{0.369\linewidth}p{0.066\linewidth}p{0.515\linewidth}}
\toprule
Parameter & Value & Role \\
\midrule
\texttt{lambda\_core\_anchor} & 0.0 & core anchors never decay \\
\texttt{lambda\_active\_context} & 0.05 & base decay rate per turn \\
\texttt{lambda\_ephemeral} & 0.69 & base decay rate per turn \\
\texttt{eta} & 0.15 & consolidation constant \\
\texttt{beta} & 0.0 & clustering protection, ablation control \\
\texttt{w\_prune} & 0.1 & demotion threshold \\
\texttt{gamma} & 0.1 & suppression multiplier \\
\texttt{max\_hops} & 2 & graph expansion depth \\
\texttt{always\_search\_cold} & true & archive searched on every query \\
\texttt{cold\_semantic\_fallback} & true & archive searched by meaning and links \\
\texttt{rrf\_k} & 60 & fusion constant \\
\texttt{rerank\_top\_k} & 20 & facts surviving into the prompt \\
\texttt{hydrate\_top\_k} & 5 & source turns attached \\
\texttt{recency\_boost} & 0.0 & ablation control \\
\texttt{weight\_signal} & 0.0 & see Section 5.6 \\
\texttt{source\_grounding} & 0.0 & see Section 5.4 \\
\texttt{compact\_validity} & true & shorter validity rendering \\
\texttt{defer\_reinforcement\_writes} & true & buffered write-back \\
\texttt{reinforcement\_buffer\_limit} & 256 & flush trigger \\
\texttt{auto\_consolidate\_turns} & 5 & consolidation interval \\
\texttt{resolve\_link\_targets} & true & link targets matched to real entities \\
\texttt{link\_match\_threshold} & 0.5 & token overlap for a link match \\
\bottomrule
\end{tabular}
\end{table}

\begin{figure*}[tb]
\centering
\includegraphics[width=0.70\textwidth]{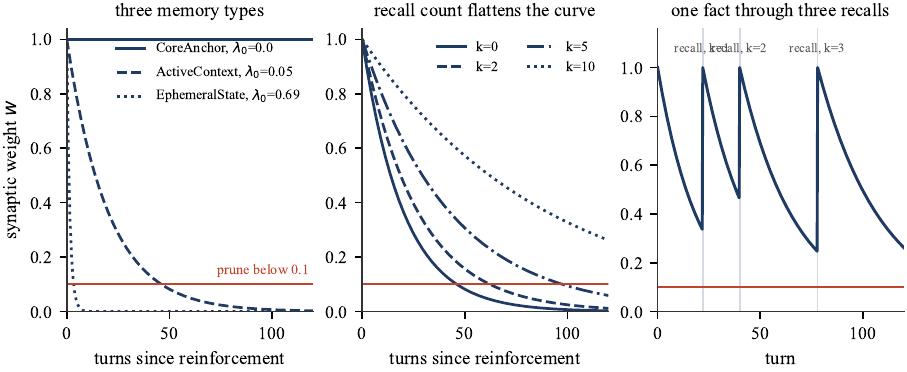}
\caption{Decay under the shipped constants. Left, the three memory types at $k = 0$. Centre, the effect of retrieval count on an active-context fact. Right, one fact through three retrievals; each resets the weight to 1 and lowers the rate for the next interval.}\label{figure:decay}
\end{figure*}

\section*{Appendix B. Statistical detail}

Under a normal approximation with a discordance rate $d$, the smallest lead detectable at $\alpha = 0.05$ on $n$ paired questions is about

\begin{equation}\label{equation:mde}
\delta_{\min} \approx \frac{1.96 \sqrt{d n}}{n}
\end{equation}

\begin{table}[tb]
\centering
\caption{Smallest lead detectable by a two-sided paired test at $\alpha = 0.05$, at the discordance rate of 0.25 observed here.}\label{table:mdetab}
\small
\begin{tabular}{p{0.258\linewidth}p{0.306\linewidth}p{0.386\linewidth}}
\toprule
Paired questions & Expected discordant & Smallest detectable lead \\
\midrule
108 & 27 & 9.4 points \\
214 & 54 & 6.7 points \\
322 & 80 & 5.5 points \\
830 & 208 & 3.4 points \\
1,539 & 385 & 2.5 points \\
1,985 & 496 & 2.2 points \\
\bottomrule
\end{tabular}
\end{table}

\begin{figure*}[tb]
\centering
\includegraphics[width=0.70\textwidth]{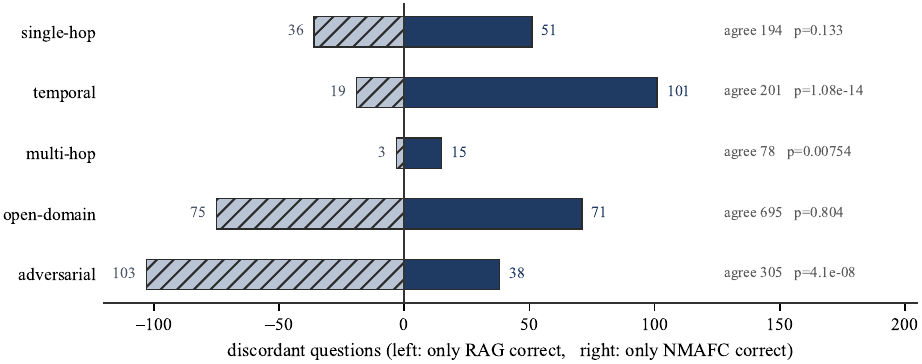}
\caption{Discordant questions per category. Bars to the right are questions only NMAFC answered correctly, bars to the left only the baseline. Agreement counts and exact p-values are on the right. The information in the agreements is zero, which is the argument for pairing.}\label{figure:disc}
\end{figure*}

Table \ref{table:byconv} breaks the paired result down by source conversation. The point of the table is the spread rather than any one row: our accuracy runs across 17 points and the baseline's across 16, and the sign of the lead changes three times. A study conducted on one conversation could have reported +9.5 or -3.1 from this same system.

\begin{table}[tb]
\centering
\caption{Paired accuracy by conversation, all five categories. Won and lost are the discordant counts, and p is McNemar exact. No single conversation is significant after any correction for ten comparisons, which is the expected result at these sample sizes given the minimum detectable effects above.}\label{table:byconv}
\scriptsize
\begin{tabular}{lrrrrrr}
\toprule
Conversation & n & NMAFC & RAG & Lead & Won/lost & p \\
\midrule
conv-26 & 199 & 68.3 & 58.8 & +9.5 & 42/23 & 0.025 \\
conv-30 & 105 & 57.1 & 55.2 & +1.9 & 16/14 & 0.856 \\
conv-41 & 193 & 66.3 & 60.6 & +5.7 & 33/22 & 0.177 \\
conv-42 & 259 & 56.4 & 59.5 & -3.1 & 30/38 & 0.396 \\
conv-43 & 242 & 60.3 & 61.6 & -1.2 & 27/30 & 0.791 \\
conv-44 & 158 & 64.6 & 58.9 & +5.7 & 25/16 & 0.211 \\
conv-47 & 190 & 73.2 & 71.1 & +2.1 & 23/19 & 0.644 \\
conv-48 & 239 & 63.2 & 64.0 & -0.8 & 27/29 & 0.894 \\
conv-49 & 196 & 60.2 & 58.7 & +1.5 & 25/22 & 0.771 \\
conv-50 & 204 & 65.2 & 62.7 & +2.5 & 28/23 & 0.576 \\
\bottomrule
\end{tabular}
\end{table}

\section*{Appendix C. Accuracy, context and latency}

\begin{figure*}[tb]
\centering
\includegraphics[width=0.70\textwidth]{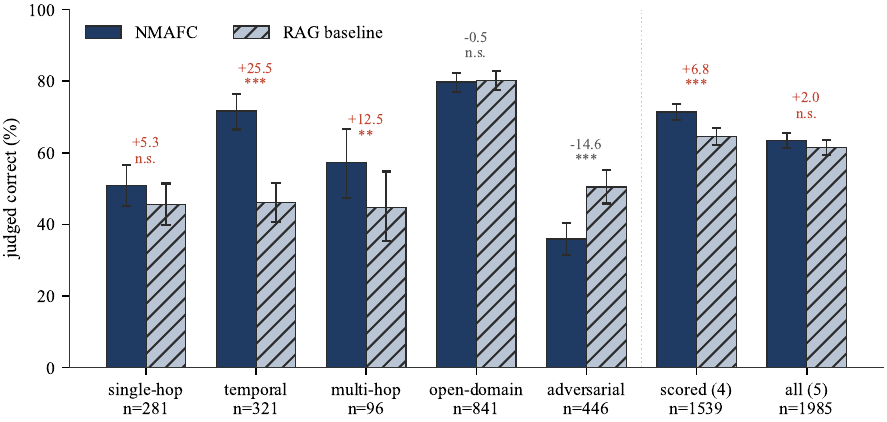}
\caption{Accuracy by category with Wilson 95 per cent intervals, plotting Table \ref{table:main}. Significance markers are McNemar exact on the paired discordant counts, so they do not follow from whether the intervals overlap.}\label{figure:acc}
\end{figure*}

\begin{figure*}[tb]
\centering
\includegraphics[width=0.70\textwidth]{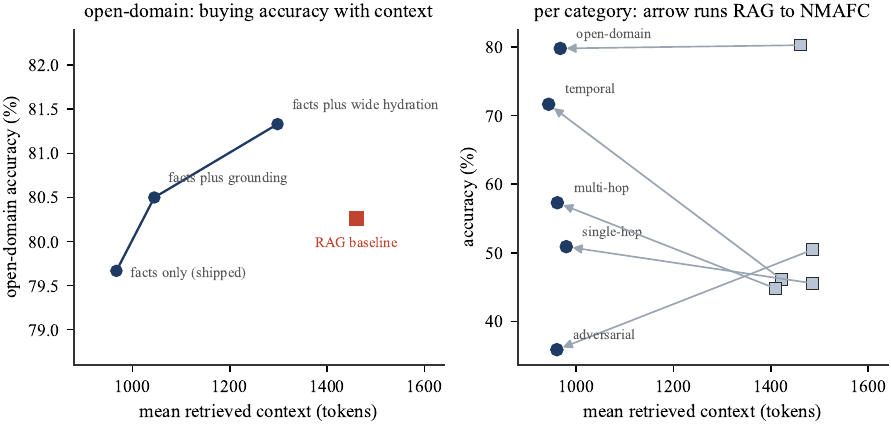}
\caption{Left, the open-domain accuracy and context of three retrieval widths against the baseline. Right, all five categories; each arrow runs from the baseline's position to ours, so an arrow pointing up and to the left is a win on both axes.}\label{figure:pareto}
\end{figure*}

\begin{figure*}[tb]
\centering
\includegraphics[width=0.70\textwidth]{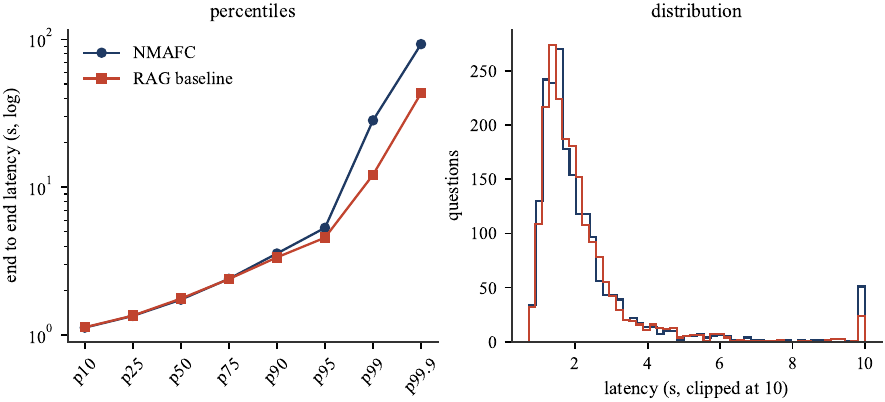}
\caption{Latency percentiles on a log scale, and the distribution clipped at ten seconds. The arms separate only above the ninetieth percentile. Both maxima near 93.8 seconds are the provider's retry ladder rather than either system's work.}\label{figure:lat}
\end{figure*}

\section*{Appendix D. Discordant questions, verbatim}

The questions below are taken from the released per-question data without editing. They are the first cases the filter returned in each direction, not a selection made after reading them. The filter for Table \ref{table:advloss} was: category adversarial, baseline correct, our arm incorrect, reference answer six words or longer. The filter for Table \ref{table:temwin} was: category temporal, our arm correct, baseline incorrect.

\begin{table*}[tb]
\centering
\caption{Adversarial questions the baseline answered and we did not. Rows one to four show wording drift, where our answer is defensible and shares almost nothing with the reference. Rows five and six show the other failure, where retrieval missed and our arm declined.}\label{table:advloss}
\small
\begin{tabular}{p{0.210\linewidth}p{0.319\linewidth}p{0.199\linewidth}p{0.222\linewidth}}
\toprule
Question & Reference & NMAFC & RAG \\
\midrule
What is Melanie excited about in her adoption process? & creating a family for kids who need one & inspired by her friend's positive experience & making a family for kids who need one \\
What did Caroline and her family do while camping? & explored nature, roasted marshmallows, and went on a hike & Caroline did not go camping with her family & explored nature, roasted marshmallows, hiked \\
What motivated Melanie to pursue counseling? & her own journey and the support she received, and how counseling improved her life & Melanie is not pursuing counseling & her own journey and the support she got \\
What inspired Melanie's painting for the art show? & visiting an LGBTQ center and wanting to capture unity and strength & nature and sunsets & the LGBTQ center visit \\
How did Caroline's son handle the accident? & He was scared but reassured by his family & Caroline does not have a son mentioned in the facts & He was scared but reassured and explained he would be OK. \\
What does Jon's tattoo symbolize? & Freedom and expressing himself through dance & No information available & freedom, dancing without worrying what people think \\
\bottomrule
\end{tabular}
\end{table*}

\begin{table*}[tb]
\centering
\caption{Temporal questions we answered and the baseline did not. The baseline's answers are frequently the relative phrase from the retrieved chunk, unresolved.}\label{table:temwin}
\small
\begin{tabular}{p{0.329\linewidth}p{0.197\linewidth}p{0.205\linewidth}p{0.219\linewidth}}
\toprule
Question & Reference & NMAFC & RAG \\
\midrule
When did Melanie paint a sunrise? & 2022 & 2022 & last year \\
When is Melanie planning on going camping? & June 2023 & June 2023 & next month \\
When did Caroline give a speech at a school? & The week before 9 June 2023 & early June 2023 & the week before 25 May 2023 \\
How long ago was Caroline's 18th birthday? & 10 years ago & ten years ago & Not mentioned in the excerpts. \\
When did Caroline have a picnic? & The week before 6 July 2023 & late June or early July 2023 & yesterday \\
When did Caroline go to the LGBTQ conference? & 10 July 2023 & 10 July 2023 & two days ago on 4 July 2023 \\
\bottomrule
\end{tabular}
\end{table*}

\section*{Appendix E. Negative and unproven results}

The answer-type gate of Section 3.7 was built on a measurement taken before it existed. Over 1,535 paired questions, the 238 that name a type scored 56.7 per cent against the baseline's 61.8, while the other 1,297 scored 66.7 against 64.9, so the deficit sat in exactly the questions the gate detects. It ships on that evidence. A clean paired A/B on the gate alone, sized to resolve two points, has not been run, which is why Section 3.7 claims nothing for it.

\begin{table*}[tb]
\centering
\caption{Three of the four mechanisms of Section 5.6, with the paired counts behind each. Won and lost are discordant questions counted for the variant against the shipped setting, and p is McNemar exact. The weight signal is excluded because it was measured as answer reachability rather than as a paired accuracy test.}\label{table:neg}
\scriptsize
\begin{tabular}{lrrrrr}
\toprule
Mechanism & n & Variant & Shipped & Won/lost & p \\
\midrule
list-shape gate, wide retrieval & 214 & 36.4 & 36.9 & 9/10 & 1.0 \\
short-phrase clause, all & 1,984 & 63.4 & 63.4 & 72/72 & 1.0 \\
short-phrase clause, adversarial & 445 & 41.1 & 35.7 & 41/17 & 0.0022 \\
short-phrase clause, temporal & 321 & 67.6 & 71.7 & 2/15 & 0.0024 \\
exact-wording clause, both & 767 & 49.5 & 50.8 & 30/40 & 0.282 \\
exact-wording clause, adversarial & 446 & 36.1 & 35.9 & 28/27 & 1.0 \\
exact-wording clause, temporal & 321 & 68.2 & 71.7 & 2/13 & 0.0074 \\
\bottomrule
\end{tabular}
\end{table*}

The list-shape gate cost 1,286 retrieved tokens against the shipped setting's 968 for a result the test cannot separate from zero. The two prompt clauses are the same trade in opposite directions, since each point of adversarial wording recovered gives back close to a point of temporal precision, which is why neither ships. The exact-wording run covers adversarial and temporal only, those being the two categories the trade acts on. The short-phrase run covers 1,984 of the 1,985 questions: one question was not answered in that run and is excluded from every row of this appendix; it is included in the main run of Table 1.

\section*{Appendix F. Released files}

\begin{table*}[tb]
\centering
\caption{The files released with the paper for recomputing every number and figure in it. The scripts share no state and none of them contacts a model.}\label{table:files}
\small
\begin{tabular}{p{0.206\linewidth}p{0.744\linewidth}}
\toprule
File & Contents \\
\midrule
\texttt{data/locomo\_all.json} & 1,985 rows, one per question, both arms' predictions, correctness, retrieved-context size and latency \\
\texttt{data/od\_ship.json} & 841 open-domain rows at the shipped retrieval width \\
\texttt{data/od\_ground.json} & 841 open-domain rows with the grounding term active \\
\texttt{data/od\_wide.json} & 841 open-domain rows at ten hydrated turns \\
\texttt{data/lg\_ship.json} & 214 rows where the list-shape detector fires, shipped width \\
\texttt{data/lg\_wide.json} & 214 rows where the list-shape detector fires, wide \\
\texttt{data/commit\_short\_all.json} & 1,984 rows with the short-answer clause in the prompt \\
\texttt{data/commit\_exact.json} & 767 rows with the exact-wording clause in the prompt \\
\texttt{analysis.py} & recomputes every table \\
\texttt{make\_figures.py} & redraws every figure \\
\texttt{references.py} & the bibliography as data, with resolved DOIs \\
\texttt{build.py} & renders this source to LaTeX, Word and Markdown \\
\bottomrule
\end{tabular}
\end{table*}

\end{document}